# ERCPMP-Gx: Endoscopic Image and Video Dataset for Morphological, Histopathological, and Genomic Characterization of Colorectal Polyposis

**Zahra Ghaffari [1], Massih Bahar[2], Mojgan Forootan[1], Ali Darvishi[3], Hamidreza Bolhasani[3]***

## Abstract

Hereditary polyposis syndromes can be precursor lesions to colorectal cancer and are associated with a broad spectrum of extracolonic tumors. Early identification and accurate classification of these syndromes are essential for timely diagnosis, individualized patient management, and targeted surveillance strategies for affected families. However, public endoscopic datasets are largely organized around the individual sporadic polyp, and none links the polyposis phenotype to histopathology and germline findings at the patient level. Here, we present ERCPMP-Gx, an endoscopic, histopathological, and genomic dataset developed to support the application of artificial intelligence (AI) in the recognition, characterization, and classification of colorectal polyposis. Most procedures were performed using the Olympus EVIS X1 system with white-light endoscopy (WLE), narrow-band imaging (NBI), magnifying NBI (M-NBI), and NBI with near focus modes, yielding 160 images and accompanying video clips. Approximately eighty percent of cases represent clinically and/or genetically confirmed hereditary polyposis syndromes (PG)—including familial adenomatous polyposis (FAP), Peutz-Jeghers syndrome (PJS), juvenile polyposis syndrome (JPS), and ganglioneuroma syndrome (GNS)—while the remaining twenty percent comprise non-hereditary polyps and polyp-mimicking lesions with overlapping morphological features (Non-PG), included to support differential classification. Each released record is linked, where available, to standardized endoscopic annotations, representative histopathology, and clinically reported germline findings, forming an AI-ready, patient-level annotation framework. The dataset is publicly accessible at Mendeley (https://doi.org/10.17632/nzyfc544bx.2). For the latest updates and further information, readers are referred to the DataBioX website: https://databiox.com.

**Keywords:** Colorectal Polyposis; Multimodal Dataset; Germline Genetics; Adenomatous Polyposis; Hamartomatous Polyposis; Narrow-Band Imaging; Artificial Intelligence

## 1. Background

Colorectal cancer (CRC) is among the leading causes of cancer-related mortality worldwide, ranking as the third most common malignancy and the second leading cause of cancer death, with an estimated lifetime risk of approximately 4–5% [1]. Hereditary polyposis syndromes account for approximately 1% of all CRC cases but carry a disproportionately high individual and familial risk [2]. Early diagnosis and precise classification of these syndromes are crucial for initiating targeted surveillance programs that facilitate early cancer detection and inform genetic counseling for affected families [2].

Artificial intelligence (AI) has shown increasing promise in colorectal cancer diagnosis and management, from polyp detection to optical histological characterization [3,4,5]. However, the application of AI to hereditary polyposis syndromes—where accurate classification requires integrating endoscopic phenotype with histopathological and germline genetic findings—remains largely unexplored, as detailed in Section 1.1.

### 1.1 Related Resources

The principal novelty of this dataset extends beyond conventional polyposis detection during colonoscopy. Instead, this study aims to develop a multimodal artificial intelligence (AI) model capable of predicting the underlying genetic alterations and histopathological characteristics of colorectal polyposis syndromes in real time during colonoscopy. The proposed model is trained using a comprehensive dataset integrating endoscopic images, histopathological findings, and germline genetic information from patients with hereditary colorectal polyposis syndromes. Compared with the existing literature, this approach represents a significant advancement, as the majority of published AI studies have primarily focused on polyp detection or optical histological characterization rather than predicting the underlying molecular and genetic background of colorectal polyposis. A comparative summary of these resources is provided in Table 1.

[1] Shahid Beheshti University of Medical Sciences, Tehran, Iran.
[2] Familial & Hereditary Cancers Institute, Tehran, Iran.
[3] Shiraz University of Medical Sciences, Shiraz, Iran.
[3] DataBioX Research, Tehran, Iran.
* hamidreza@databiox.com

**Table 1. Comparison of ERCPMP-Gx with Existing Colorectal Polyp/Polyposis AI Studies and Datasets**

| Study | Year | Sample Size | Data / Imaging | AI Purpose | Histopathology | Hereditary Focus | Real-time AI | Target Disease | Main Contribution | Main Limitation |
|---|---|---|---|---|---|---|---|---|---|---|
| Bhandari et al. [6] | 2017 | 40 | Colonoscopy images + clinical variables | Standardized clinical dataset | Yes | No | No | Large non-pedunculated colorectal polyps | First evidence-based standardized dataset for colorectal polyp assessment | No AI, no genomic integration, limited sample size |
| Bern et al. [7] | 2019 | 125 | NBI colonoscopy | Deep neural network for optical biopsy | Yes | No | Yes | Sporadic diminutive polyps | Accurate real-time differentiation of adenomatous and hyperplastic polyps | Binary classification only; no genomic or patient-level information |
| Borgli et al. (HyperKvasir) [8] | 2020 | 110,079 images + 374 videos | GI endoscopy | Detection, classification and segmentation dataset | Partial | No | Potential | Gastrointestinal lesions | Largest publicly available GI endoscopy dataset | Imaging not linked to pathology or genomics |
| AI Diagnostic Accuracy Meta-analysis [9] | 2021 | Multiple studies | Colonoscopy | Evaluation of AI diagnostic performance | Yes | No | Variable | Sporadic colorectal polyps | Demonstrated high diagnostic accuracy of AI-assisted optical biopsy | No multimodal learning or genomic prediction |
| Parsa & Byrne [10] | 2021 | Review | Colonoscopy | Review of CADe/CADx systems | Yes | No | Variable | General colorectal polyps | Comprehensive review of AI applications in colonoscopy | No original dataset or multimodal framework |
| Picon et al. (PICCOLO HE/MPM) [11] | 2022 | 50 | Histopathology imaging | Tissue classification and virtual staining | Yes | No | No | Colorectal adenoma and adenocarcinoma | Introduced a multimodal pathology imaging dataset | Ex vivo imaging only; no colonoscopy or genomic information |
| Keshtkar et al. [12] | 2023 | Multiple studies | Colonoscopy | CNN-based diagnosis | Yes | No | Variable | Sporadic colorectal polyps | Demonstrated effectiveness of CNN-based diagnosis | Image-based diagnosis only |

| Foroutan et al. (ERCPMP) [13] | 2024 | 217 | White-light colonoscopy | Detection, classification and segmentation dataset | Yes | No | Potential | Colorectal polyps | Combined imaging, morphology and pathology using Paris, Pit Pattern and JNET classifications | No hereditary polyposis cases or germline genomic data |
|---|---|---|---|---|---|---|---|---|---|---|
| Nie et al. [14] | 2024 | Review | Colonoscopy | Review of real-time AI systems | Yes | No | Yes | Sporadic colorectal polyps | Summarized current AI technologies in colonoscopy | No original multimodal framework or genomic integration |
| Soleymanjahi et al. [15] | 2024 | Multiple RCTs | Colonoscopy | AI-assisted adenoma detection | Yes | No | Yes | General colorectal polyps | Demonstrated improved adenoma detection rate | Focused only on CADe performance |
| Proposed Study | Present | Colorectal polyposis patients | Integrated colonoscopy, histopathology and germline genomics | Multimodal deep learning for simultaneous pathology and genotype prediction | Yes | Yes | Yes | Colorectal polyposis syndromes | First real-time multimodal AI framework integrating colonoscopic phenotype, histopathology and germline genomics for simultaneous pathological classification and mutation prediction | External multicenter validation required before clinical implementation |

Abbreviations: CNN, convolutional neural network; DNN, deep neural network; NBI, narrow-band imaging; CADe, computer-aided detection; CADx, computer-aided diagnosis.

## 1.2 Novelty

Beyond the general application of artificial intelligence to colorectal polyposis detection, the specific novelty of ERCPMP-Gx lies in its patient-level integration of three previously disconnected data modalities—colonoscopic imaging across multiple modes (WLE, NBI, M-NBI), histopathological confirmation, and germline genetic results—within a single, structured, AI-ready annotation framework. This structure enables, for the first time in a publicly available resource, the joint modeling of endoscopic phenotype and underlying genotype for hereditary polyposis syndromes, rather than treating these as separate downstream analyses.

Prospective studies with larger, multicenter cohorts—including additional images, videos, and longitudinal clinical follow-up—will be required to validate the clinical utility of AI models trained on this dataset. Such efforts could meaningfully contribute to earlier diagnosis, more accurate syndromic classification, and personalized surveillance strategies for patients and families affected by hereditary colorectal polyposis.

# 2. Method

This applied research study was conducted among consecutive patients diagnosed with hereditary colorectal polyposis syndromes and the non-hereditary polyps and polyp-mimicking lesions with overlapping morphological features (Non-PG) who were referred from the FamCan (Familial and Hereditary Cancer Institute) to the advanced Gastroenterology ward. These procedures were captured by the Olympus EVIS X1 system with white-light endoscopy (WLE), narrow-band imaging (NBI) and magnifying NBI (M-NBI), and NBI with near focus modes. The colonoscopic findings were integrated into the dataset following confirmation by the corresponding histopathological and genetic findings. The gathered images and videos after pre-processing, are converted to PNG, and MP4 formats and presented in the dataset. The study was carried out between 2024 and 2026.

**Inclusion criteria:** Patients were eligible if they had a clinically and/or genetically confirmed diagnosis of a hereditary colorectal polyposis syndrome (FAP, JPS, PJS, GNS, or Cowden syndrome), were referred from the FamCan clinic to the advanced Gastroenterology ward between 2024 and 2026, and provided informed consent for anonymized data sharing. Patients presenting with lesions clinically mimicking polyposis (e.g., inflammatory like IBD, P-SRUS polyps, and Pneumatosis Intestinalis [PI]), similarly referred from the FamCan clinic, were also included to support differential-diagnosis modeling.

**Exclusion criteria:** Cases were excluded if endoscopic, histopathological, or genetic data were incomplete, if image or video quality was insufficient for annotation, or if informed consent was not obtained. Excluded cases were documented separately and are not included in the released dataset (see Section 5.1, Quality Control and Ground-Truth Verification).

Each participant was assigned to an endoscopic phenotype category based on colonoscopic findings, and separately to a diagnostic outcome describing whether a hereditary polyposis syndrome was genetically and/or clinically confirmed (PG) or not (Non-PG). Separating the endoscopic phenotype from the diagnostic outcome avoids the circularity of labeling an image with a diagnosis that could not have been known at the time of the examination, and reflects the intended real-world use case in which the endoscopic phenotype is observed before the underlying genotype is known.

All procedures were performed using the Olympus EVIS X1 system (processor: EVIS EXTRA III CV-1500, or CV-190; Olympus Corp., Tokyo, Japan). In most cases, a CF-EZ1500DI/L colonoscope was used. Imaging modalities included white-light endoscopy (WLE), magnifying narrow-band imaging (M-NBI), and narrow-band imaging (NBI) with near focus.

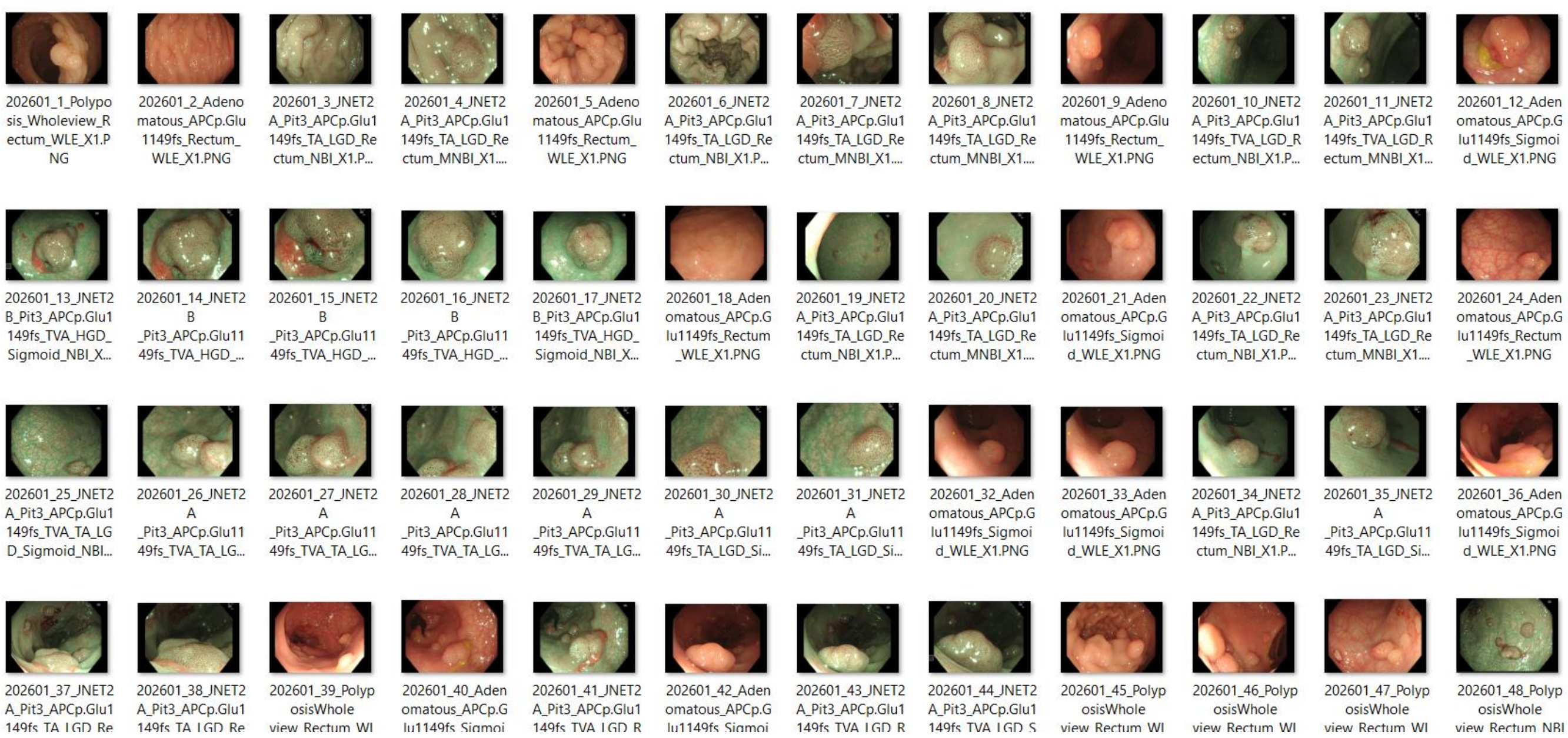

**Figure 1.** Representative colonoscopic images from the ERCPMP-Gx dataset. The file naming convention encodes key annotations, including imaging modality, anatomical location, histopathology, and Genetics.

All extracted colonoscopic, histopathological, and genetic information for each case was subsequently entered into a structured Excel (xlsx) file as standardized metadata, linked to the corresponding image and video files via unique case identifiers.

| Patient Demographic & General Information | | | | | | Chief Complaint &History | | Anatomical Features | | Pathology | Genetics | |
|---|---|---|---|---|---|---|---|---|---|---|---|---|
| Patient Code | Procedure Year | Endoscopy Processor | Endoscopy Device | Sex | Age | Reason for the Procedure | Family History | Polyp Location | Mucus Filling | Dysplasia grade & differentiation | Gene | Comment |
| 202601 | 2024 | EVIS X1 | CF-EZ15ooDI/L | Female | 23 | APC mutation | Positive | Rectum | Negative | LGD | APC mutation | AFAP |
| 202601 | 2024 | EVIS X1 | CF-EZ15ooDI/L | Female | 23 | APC mutation | Positive | Rectum | Negative | LGD | APC mutation | AFAP |
| 202601 | 2024 | EVIS X1 | CF-EZ15ooDI/L | Female | 23 | APC mutation | Positive | Rectum | Negative | LGD | APC mutation | AFAP |
| 202601 | 2024 | EVIS X1 | CF-EZ15ooDI/L | Female | 23 | APC mutation | Positive | Rectum | Negative | LGD | APC mutation | AFAP |
| 202601 | 2024 | EVIS X1 | CF-EZ15ooDI/L | Female | 23 | APC mutation | Positive | Rectum | Negative | LGD | APC mutation | AFAP |
| 202601 | 2024 | EVIS X1 | CF-EZ15ooDI/L | Female | 23 | APC mutation | Positive | Rectum | Negative | LGD | APC mutation | AFAP |
| 202601 | 2024 | EVIS X1 | CF-EZ15ooDI/L | Female | 23 | APC mutation | Positive | Rectum | Negative | LGD | APC mutation | AFAP |
| 202601 | 2024 | EVIS X1 | CF-EZ15ooDI/L | Female | 23 | APC mutation | Positive | Rectum | Negative | LGD | APC mutation | AFAP |
| 202601 | 2024 | EVIS X1 | CF-EZ15ooDI/L | Female | 23 | APC mutation | Positive | Rectum | Negative | LGD | APC mutation | AFAP |
| 202601 | 2024 | EVIS X1 | CF-EZ15ooDI/L | Female | 23 | APC mutation | Positive | Rectum | Negative | LGD | APC mutation | AFAP |
| 202601 | 2024 | EVIS X1 | CF-EZ15ooDI/L | Female | 23 | APC mutation | Positive | Rectum | Negative | LGD | APC mutation | AFAP |
| 202601 | 2024 | EVIS X1 | CF-EZ15ooDI/L | Female | 23 | APC mutation | Positive | Rectum | Negative | LGD | APC mutation | AFAP |
| 202601 | 2024 | EVIS X1 | CF-EZ15ooDI/L | Female | 23 | APC mutation | Positive | Rectum | Negative | LGD | APC mutation | AFAP |
| 202601 | 2024 | EVIS X1 | CF-EZ15ooDI/L | Female | 23 | APC mutation | Positive | Rectum | Negative | LGD | APC mutation | AFAP |
| 202601 | 2024 | EVIS X1 | CF-EZ15ooDI/L | Female | 23 | APC mutation | Positive | Rectum | Negative | LGD | APC mutation | AFAP |
| 202601 | 2024 | EVIS X1 | CF-EZ15ooDI/L | Female | 23 | APC mutation | Positive | Rectum | Negative | LGD | APC mutation | AFAP |
| 202601 | 2024 | EVIS X1 | CF-EZ15ooDI/L | Female | 23 | APC mutation | Positive | Sigmoid | Negative | LGD | APC mutation | AFAP |
| 202601 | 2024 | EVIS X1 | CF-EZ15ooDI/L | Female | 23 | APC mutation | Positive | Sigmoid | Negative | LGD | APC mutation | AFAP |
| 202601 | 2024 | EVIS X1 | CF-EZ15ooDI/L | Female | 23 | APC mutation | Positive | Sigmoid | Negative | LGD | APC mutation | AFAP |
| 202601 | 2024 | EVIS X1 | CF-EZ15ooDI/L | Female | 23 | APC mutation | Positive | Sigmoid | Negative | LGD | APC mutation | AFAP |
| 202601 | 2024 | EVIS X1 | CF-EZ15ooDI/L | Female | 23 | APC mutation | Positive | Sigmoid | Negative | LGD | APC mutation | AFAP |
| 202601 | 2024 | EVIS X1 | CF-EZ15ooDI/L | Female | 23 | APC mutation | Positive | Sigmoid | Negative | LGD | APC mutation | AFAP |

**Figure 2.** Example of the structured annotation and metadata (XLSX) file used in the ERCPMP-Gx dataset, showing per-case fields for demographic information, chief complaint and family history, anatomical features, histopathological grading, and germline genetic findings.

Still images were captured in JPG and PNG formats at 2879×1799 pixels (standard full-frame) and 657×454 pixels (magnified near-focus view). Video clips (Full video, 1920×1080 resolution) were obtained from relevant anatomical locations for each patient (rectum, left colon, transverse colon, right colon). For each identified lesion, morphological characteristics were assessed according to internationally accepted endoscopic classification systems. The Paris Classification was applied to all polypoid lesions. In addition, Pit Pattern and JNET classifications were recorded for adenomatous, hamartomatous, serrated, and other relevant polyps whenever applicable. Additional lesion characteristics, including polyp size and number, were documented whenever feasible. Representative biopsy specimens were obtained from the largest lesion within each morphologically distinct polyp category, based on the clinical judgment of the endoscopist. Following proper orientation, tissue samples were submitted for histopathological examination. All cases were confirmed by histopathological examination. Genetic, endoscopic, and histopathological findings were integrated into the ERCPMP-Gx dataset. An overview of this multimodal data structure is presented in Fig. 3.

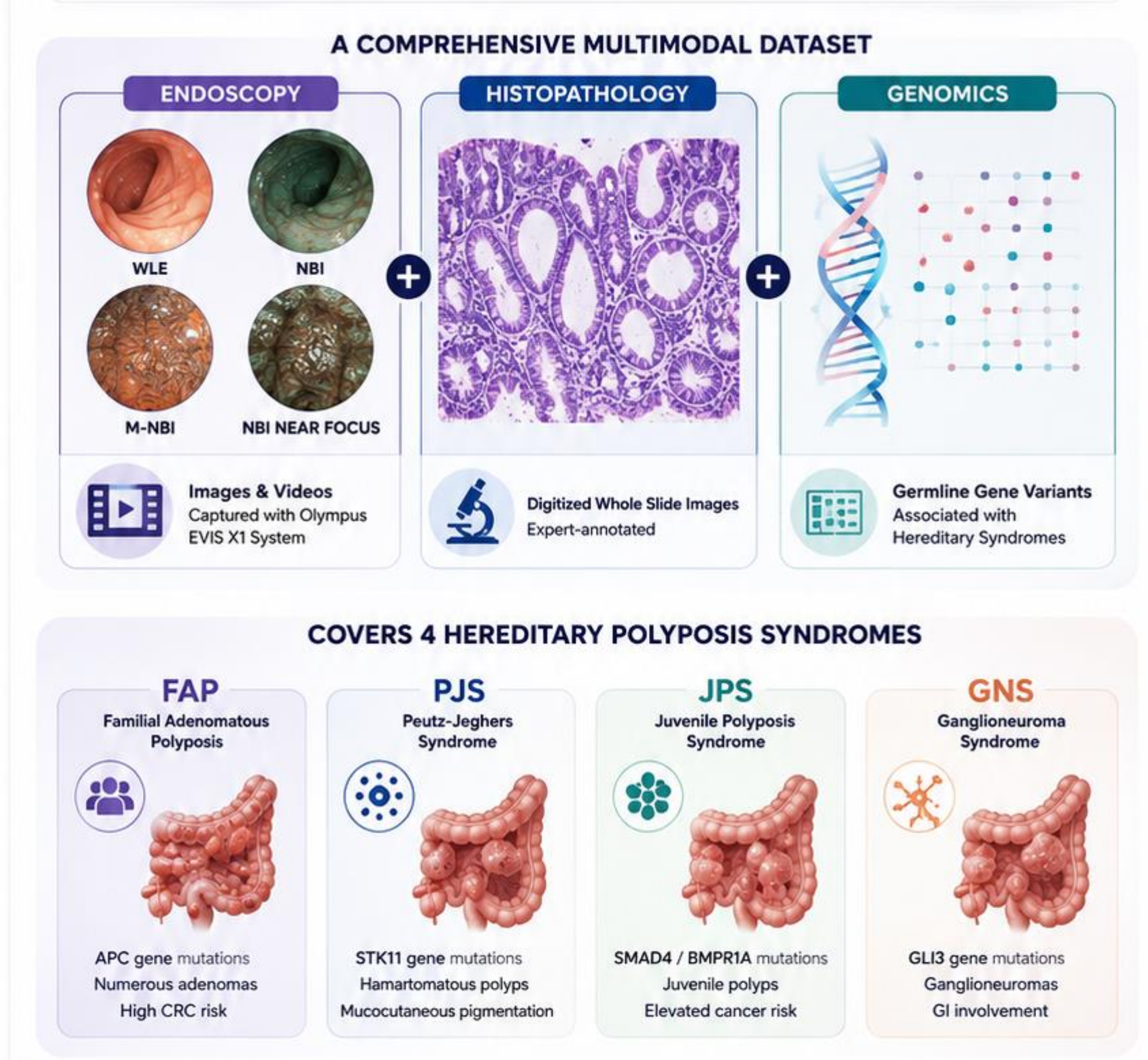

**Figure 3.** Graphical abstract of the ERCPMP-Gx dataset, illustrating the integration of multimodal endoscopic imaging, histopathology, and germline genomics across four hereditary polyposis syndromes (FAP, PJS, JPS, and GNS).

## 3. Data Records and Analysis

Demographic and pathological characteristics are summarized in Table 2.

**Table 2.** Demographic and pathological characteristics of the study population.

| Characteristic | FAP | JPS | PJS | GNS | Inflammatory | PI | Cowden |
|---|---|---|---|---|---|---|---|
| Image Number | 56 | 40 | – | 40 | – | – | – |
| Pathology | Adenomatous | Hamartomatous | Hamartomatous | Hamartomatous | – | – | Hamartomatous |
| Genetics | APC mutation | SMAD4/BMPR 1A | STK11 | – | – | – | PTEN |

# 4. Validation

## 4.1 Quality Control and Ground-Truth Verification

All image and video files were inspected for completeness, correct anatomical and genetic labeling, and absence of identifiable information prior to inclusion. Ground truth labels were established through a rigorous four-step process: (i) comprehensive colonoscopic examination with targeted biopsy or polypectomy of representative lesions; (ii) independent histopathological assessment by two board-certified gastrointestinal pathologists; (iii) expert annotation of endoscopic and morphological findings by multiple experienced gastroenterologists with advanced colonoscopy expertise; and (iv) consensus-based adjudication of all discordant cases.

In addition to endoscopic and histopathological findings, genetic data and detailed morphological characteristics of each lesion were incorporated into the reference standard to provide high-quality ground truth for training and validating the artificial intelligence model in the diagnosis and classification of polyposis syndromes.

Cases with incomplete clinical, endoscopic, pathological, or genetic data were excluded from the final reference dataset and were appropriately documented in the annotation file.

For every lesion, the metadata explicitly distinguish whether the recorded diagnosis reflects histopathological confirmation or an endoscopist's optical assessment. This distinction is maintained throughout the dataset so that users are not led to treat optical impressions as a histopathological reference standard.

## 4.2 Annotation Consistency

The annotation of colonoscopic findings was performed by multiple experienced gastroenterologists using standardized criteria that were applied consistently throughout the dataset. The collected clinical, morphological, pathological, and genetic information from patients with polyposis was integrated to develop and train artificial intelligence models for the detection and characterization of polyposis-related lesions. Formal inter-rater and intra-rater reliability assessments were not conducted in this pilot study, which represents a recognized limitation. Users should be aware that certain morphological and pathological features may be subject to inter-observer variability despite the use of standardized annotation protocols.

## 4.3 Benchmark Protocol

No predefined training, validation, or test split was established for this study. Since the primary objective is to develop an artificial intelligence model for the diagnosis and classification of polyposis using colonoscopic findings, morphological characteristics, histopathological features, and genetic data, dataset partitioning should be performed according to the study design. To prevent data leakage and ensure robust model generalizability, patient-level data splitting is recommended, ensuring that all data from a single patient are assigned exclusively to either the training, validation, or test set.

The dataset supports several clinically relevant artificial intelligence tasks, including the detection of polyposis, classification of polyp subtypes based on morphological and histopathological characteristics, prediction of genetically associated polyposis syndromes, and risk stratification for malignant transformation. Model performance should be evaluated using appropriate metrics, such as the area under the receiver operating characteristic curve (AUROC), sensitivity, specificity, accuracy, positive predictive value (PPV), negative predictive value (NPV), and F1-score, depending on the intended clinical application. For screening purposes, maximizing sensitivity is essential to minimize missed lesions, whereas for diagnostic applications, an optimal balance between sensitivity and specificity should be achieved.

Given the current dataset size, this resource is well suited for exploratory analyses, pilot studies, and proof-of-concept model development. Ongoing efforts are focused on expanding the dataset with additional cases, images, and video data, which will enable more comprehensive benchmarking and robust model validation in future releases.

# 5. Discussion

## 5.1 Strengths

In this study, we present ERCPMP-Gx as a clinically grounded and systematically annotated dataset designed to accelerate reproducible artificial intelligence (AI) research in polyposis syndromes. We aimed to introduce a multimodal artificial intelligence framework from patients with colorectal polyposis syndromes by integrating high-quality endoscopic imaging, histopathological and genetical confirmation, and morphological metadata, Current public datasets generally lack comprehensive patient-level integration of endoscopic findings with histopathological, morphological, and germline genetic information. Consequently, existing AI systems are capable of identifying or classifying colorectal polyps but are unable to infer the underlying hereditary syndrome or predict disease-causing genetic mutations.

Unlike previous image-based computer-aided detection (CADe) and computer-aided diagnosis (CADx) systems, the proposed framework aims to perform real-time prediction of both histopathological and morphological characteristics and the underlying genotype, thereby facilitating precision endoscopy, individualized surveillance, and personalized clinical decision-making.

A key strength of this study lies in the use of multimodal endoscopic data, including both still images and video clips acquired under standardized imaging modalities (WLE, M-NBI, and NBI with near focus). Furthermore, all cases were confirmed by histopathological examination, ensuring high diagnostic validity and minimizing labeling errors.

Another important strength of this dataset is the comprehensive integration of genetic information, colonoscopic findings, morphological characteristics of polyps, and histopathological diagnoses collected from patients with polyposis. Patients with a clinically and/or genetically confirmed hereditary polyposis syndrome were referred for colonoscopic examination, during which lesion assessments were performed by experienced gastroenterologists, while histopathological evaluation provided complementary diagnostic confirmation. This multimodal approach enables the development of interpretable artificial intelligence models that more closely reflect real-world clinical decision-making in the diagnosis and management of polyposis syndromes.

Importantly, the study design mirrors routine clinical practice. Patients with a clinically and/or genetically confirmed hereditary polyposis syndrome were referred for colonoscopic examination, during which polyps were identified and characterized according to their morphological features. Suspicious lesions were subsequently resected or biopsied for histopathological confirmation. The integration of genetic, endoscopic, and pathological data enhances the clinical value of the dataset and provides a robust foundation for developing AI models capable of improving the detection, classification, and risk stratification of colorectal polyps and polyposis syndromes.

## 5.2 Limitations and Biases

The primary limitation of this study is the relatively limited number of patients with hereditary colorectal polyposis syndromes, reflecting the rarity of these disorders in clinical practice. Although all cases were confirmed by histopathological examination and germline genetic testing, larger multicenter cohorts will be required to improve the robustness and generalizability of future AI models.

Another limitation is the imbalance among different polyposis syndromes and pathogenic variants. More prevalent conditions are represented more frequently than rare hereditary syndromes, which may introduce bias during model training. Furthermore, variations in image quality, endoscopic equipment, and acquisition protocols may contribute to data heterogeneity.

Finally, the current dataset represents patients referred to specialized centers and may not fully reflect the characteristics of general screening populations. Future work will focus on expanding the cohort through multicenter collaboration, increasing the representation of rare genetic subtypes, and incorporating longitudinal clinical follow-up to further improve the clinical applicability of multimodal AI models.

## 5.3 Responsible Use

The proposed dataset is intended exclusively for non-commercial scientific research on hereditary colorectal polyposis syndromes and related colorectal neoplasia. It is designed to support the development and evaluation of artificial intelligence algorithms for multimodal learning, including colonoscopic image analysis, histopathological classification, genotype prediction, phenotype-genotype correlation, and clinical decision-support research.
All patient data have been fully anonymized before inclusion in the dataset. Users must not attempt to re-identify any participant or combine the dataset with external sources for patient identification.
Models developed using this dataset should not be implemented in routine clinical practice without rigorous external validation, prospective multicenter evaluation, and regulatory approval. Because the dataset represents patients referred to specialized hereditary gastrointestinal cancer centers, disease prevalence, mutation spectrum, and phenotypic characteristics may differ substantially from those observed in general screening populations. Consequently, recalibration and independent validation are recommended before deployment in other clinical settings.
Researchers should also acknowledge the relatively limited representation of rare hereditary polyposis syndromes and uncommon pathogenic variants when interpreting model performance. Potential biases arising from class imbalance, referral patterns, and imaging heterogeneity should be carefully considered in downstream analyses and publications.
The dataset is intended solely for research and educational purposes and must not be used as the sole basis for clinical diagnosis, surveillance recommendations, therapeutic decision-making, or genetic counseling.

## 6. Conclusion

Despite these strengths, several limitations should be acknowledged. First, the sample size may be relatively limited for training highly complex deep learning models, particularly for the classification of rare hereditary polyposis subtypes. Second, as the study was conducted at specific centers, the diversity of patient populations and imaging conditions may be limited, potentially affecting the generalizability of the developed models. Third, interobserver variability in the assessment of colonoscopic morphology and histopathological interpretation cannot be completely excluded, highlighting the need for future studies incorporating multi-center datasets and multi-expert annotations.

Future research should focus on expanding the dataset by including larger and more diverse patient cohorts from multiple institutions. The incorporation of high-definition colonoscopy videos, advanced imaging modalities, longitudinal clinical follow-up, and additional molecular or genomic biomarkers may further enhance model performance and clinical applicability. Prospective external validation will also be essential to confirm the robustness and generalizability of the proposed AI models. Ultimately, AI systems trained on integrated colonoscopic, pathological, and genetic data have the potential to assist gastroenterologists in the early detection of polyposis syndromes, accurate characterization of colorectal polyps, personalized risk assessment, and optimization of surveillance and therapeutic strategies.

In conclusion, ERCPMP-Gx represents an important step toward the development of clinically applicable artificial intelligence tools for colorectal polyposis. With further expansion and external validation, it has the potential to improve diagnostic accuracy, facilitate personalized patient management, and support precision medicine in hereditary and sporadic colorectal polyposis.

It should be emphasized that this data descriptor does not establish that a pathogenic germline variant can be predicted accurately or safely in real time from colonoscopic imaging. Such a claim would require a prespecified modeling study with sufficiently large and representative training and external test cohorts, patient-level data separation, uncertainty calibration, and prospective clinical validation. Accordingly, real-time genotype prediction is described here only as a potential future research direction enabled by this resource, not as a demonstrated capability.

## 7. Resource Availability

### 7.1 Data Location

ERCPMP-Gx is publicly available at Mendeley dataverse via https://doi.org/10.17632/nzyfc544bx.2. The repository contains endoscopic still images (JPG and PNG, 160 files), video clips, and anonymised metadata. All annotations and metadata are provided in a structured Excel (xlsx) file.

All files are organized with consistent naming conventions, and linkage between media files and metadata is established via unique identifiers. No predefined train/validation/test split is included in the current release.

Access is open and unrestricted. For the latest updates, dataset expansions, and further information, readers are encouraged to visit the DataBioX website: https://databiox.com.

### 7.2 Use Cases and Licensing

The dataset is intended to support non-commercial research on hereditary colorectal polyposis syndromes and the development of multimodal artificial intelligence models. Potential machine learning applications include colorectal polyp detection, histopathological classification, genotype prediction, phenotype–genotype association analysis, and clinical decision-support research by integrating colonoscopic, histopathological, and genomic data. The dataset is provided solely for research and educational purposes and should not be used for direct clinical diagnosis or therapeutic decision-making without independent prospective validation. Re-identification attempts, commercial use, and use in surveillance or decision-making applications are strictly prohibited under the terms of the CC BY-NC 4.0 license (https://creativecommons.org/licenses/by-nc/4.0/).

## Funding

Not Applicable.

## Author contributions

Z.G: Conceptualization, data curation, annotation, methodology, resources, formal analysis, writing. M.F.: Data curation, colonoscopic procedures, project administration, writing, polyps' information(morphology, surface pattern, and anatomical features). H.B.: Conceptualization, methodology, software, data analysis, writing. M.B.: Genetic data, data curation, writing. All authors approved the final manuscript.

## Conflicts of Interest

The authors declare no competing financial or non-financial interests.

## Ethics Statement

This study has been approved by the ethics committee of the Gastroenterology and Liver Disease Research Center, Research Institute for Gastroenterology and Liver Diseases, Shahid Beheshti University of Medical Sciences( Ethics approval code: IR.SBMU.MSP.REC.1404.591). According to ethical principles, the datasets completely anonymous. Informed consent was obtained from all subjects and/or their legal guardian(s) in the ethical approval and consent to participate sub-section.